\documentclass[lettersize,journal]{IEEEtran}
\usepackage{amsmath,amsfonts,amsthm,amssymb}
\usepackage{algorithmic}
\usepackage{algorithm}
\usepackage{array}
\usepackage[caption=false,font=normalsize,labelfont=sf,textfont=sf]{subfig}
\usepackage{textcomp}
\usepackage{stfloats}
\usepackage{url}
\usepackage{verbatim}
\usepackage{graphicx}
\usepackage{CJKutf8}
\usepackage{tabularx}
\usepackage{graphicx}
\usepackage{caption}
\usepackage{utfsym}
\usepackage{cite}
\usepackage{lineno} 
\usepackage[backref=page,bookmarks=true]{hyperref}
\usepackage[T1]{fontenc}
\usepackage{soul, color, xcolor} 
\usepackage{ulem} 
\usepackage{color,colortbl}
\usepackage{xcolor}
\usepackage{soul}

\begin{document}
\begin{CJK}{UTF8}{gbsn}

\title{Domain-Division based Progressive Learning \\for Source-Free Domain Adaptation}

\author{
Pan Liu, 
Jing Li,~\IEEEmembership{Member,~IEEE,}
Meng Zhao,~\IEEEmembership{Member,~IEEE,}
Wanli Xue,~\IEEEmembership{Member,~IEEE,}\\
Qinghua Hu,~\IEEEmembership{Senior Member,~IEEE,}
and Shengyong Chen,~\IEEEmembership{Senior Member,~IEEE}
\thanks{Pan Liu, Jing Li, Meng Zhao, Wanli Xue, and Shengyong Chen are with the School of Computer Science and Engineering, Tianjin University of Technology, Tianjin 300384, China (jing\underline{ }li@tju.edu.cn). (\textit{Corresponding author: Jing Li.})

Qinghua Hu is with the College of Intelligence and Computing, Tianjin University, Tianjin 300350, China 
(huqinghua@tju.edu.cn).}
}

\maketitle

\begin{abstract}
With growing privacy and portability concerns, source-free domain adaptation requires only a source pre-trained model and an unlabeled target domain, allowing for effective adaptation to the target data. Most existing self-training methods focus on selecting and exploiting samples with reliable predictions, often neglecting others. Inspired by the finding that deep models learn clean samples faster than noisy ones, we propose a domain-division based progressive learning method named DPL. Specifically, our approach consists of two alternating stages, each beginning with the division of the target domain into easy-to-adapt and hard-to-adapt subdomains based on adaptation difficulty, followed by neighborhood-based pseudo label assignment. In stage one, we enhance classification accuracy through uncertainty-aware self-training and alignment of corresponding classes between subdomains. Stage two then applies tailored learning strategies to each subdomain, starting with consistency learning on the easy-to-adapt samples and progressing to utilizing local structural information for the more challenging ones, thereby mining the intrinsic properties of the target data. Extensive experiments on several widely used benchmarks validate the effectiveness of our approach, demonstrating superior performance compared to state-of-the-art methods. Our code is available at https://github.com/iamjingli/DPL. 
\end{abstract}


\section{Introduction}
\IEEEPARstart{D}{eep} learning has recently achieved significant success in various fields\cite{bengio2021deep}, but it remains limited by the need for extensive data annotation. To address this challenge, unsupervised domain adaptation (UDA) transfers knowledge from labeled source domains to unlabeled target domains, leading to notable advancements\cite{9802910}. A common limitation of these methods is the need to access source domain data during the domain adaptation process. In real-world applications, privacy and portability concerns often prevent access to source data, rendering UDA methods impractical. 

To minimize dependency on source data, source-free domain adaptation (SFDA)\cite{li2024comprehensive} requires only a model pre-trained on the source domain and an unlabeled target domain, adapting the model to the target data. Existing SFDA approaches can be classified into two broad categories: domain generation\cite{tian2021vdm} and self-training\cite{liang2020we}. Domain generation constructs a proxy domain for compensating the unavailable source data, then the UDA techniques can be applied. But the data generation could be computationally expensive, and it is challenging to approximate the original source data when confronted with complex underlying data patterns. As a popular self-training method, pseudo labeling\cite{  liang2020we} assigns target samples pseudo labels based on the predictions of the source pre-trained model. The pseudo labels are then utilized for the subsequent self-training. Due to domain shift, the pseudo labels are noisy, leading to accumulated prediction errors. Therefore, many strategies\cite{chu2022denoised,ahmed2022cleaning,chen2022self} are proposed to improve the accuracy of pseudo labels. Accompanying pseudo labeling, intrinsic properties of the target data can be used through neighborhood structure\cite{NAMI2024,yang2021exploiting,yang2021generalized}. Nonetheless, simple neighborhood consistency is not able to fully utilize the topological information of the target. Moreover, according to our experimental observation, the efficacy of neighborhood clustering is limited when faced with large domain gap, where drawing dissimilar neighbors may lead to negative transfer\cite{NTsurvey}.

SHOT++\cite{liang2021source} splits the target data into two parts based on the prediction confidence of the source pre-trained model, and employ semi-supervised learning\cite{berthelot2019mixmatch} to improve the classification accuracy of the less-confident part in the target domain.
ICPR\cite{10438926} generates strongly augmented views specifically for selected reliable samples and enforce prediction consistency among the weakly, the strongly augmented views of the same image, and neighboring samples. 
On one hand, although these works lay emphasis on selecting samples with reliable predictions and utilizing structure information among these samples, they leave information contained in other samples under utilization;
On the other hand, deep models are known to learn in a progressive manner, first focusing on clean samples and then on noisy ones\cite{arpit2017closer,zhang2017understanding}. 
Motivated by this, we propose first dividing the target domain into two subdomains: easy-to-adapt and hard-to-adapt. We then enhance classification accuracy through progressive learning by first aligning corresponding classes between the subdomains, followed by applying tailored learning strategies to each, starting with the easy-to-adapt subdomain and progressing to the more challenging one.    

Specifically, the adaptation process of our proposed domain-division based progressive learning (DPL) method consists of two alternating stages.  
Each stage begins with dividing the data based on the reliability of the model predictions for the target samples, followed by pseudo label assignment. In stage one, the model is optimized by uncertainty-aware self-training, during which the effect of samples with uncertain predictions is reduced. Since the predictions of easy-to-adapt samples are reliable, our proposed graph contrastive learning encourages hard-to-adapt samples to align with them. In stage two, we apply distinct strategies to the subdomains, beginning with the easy-to-adapt one and advancing to the hard-to-adapt one. For the easy-to-adapt subdomain, we use standard self-training and consistency learning in the output space, while for the hard-to-adapt subdomain, we apply instance contrastive learning based on the clustering assumption\cite{jiang2022semi}, also in the output space. 

The progressive learning in our method has two aspects. First, the transition from stage one to stage two allows us to better exploit the structure of the target data, which is underutilized in stage one. Second, within stage two, learning progresses from easy-to-adapt to hard-to-adapt samples, as the latter are more challenging.

The contributions of our work are summarized as follows:
\begin{itemize}
\item[$\bullet$] We propose an integrated method called DPL for SFDA, which first divides the target domain into easy-to-adapt and hard-to-adapt subdomains based on adaptation difficulty, then applies tailored learning strategies to each.

\item[$\bullet$] The training process of the target model alternates between two stages, each beginning with domain division and pseudo-label assignment. Stage one features uncertainty-aware self-training and subdomain alignment via graph contrastive learning. In stage two, the process advances by fully utilizing the intrinsic structure of the target data, starting with the easy-to-adapt subdomain and advancing the harder one. 

\item[$\bullet$] Extensive experiments on several widely used benchmarks verify the effectiveness of our approach, demonstrating better performance than state-of-the-art methods. 
\end{itemize}

The remainder of this paper is organized as follows. Section II introduces related works. Section III describes the problem setting and elaborates on our proposed DPL method. Experimental results on various benchmarks are presented in Section IV. Finally, we conclude our work in Section V.

\section{Related Work}
\subsection{Unsupervised Domain Adaptation}
Unsupervised domain adaptation (UDA)\cite{9802910} is a well-established transfer learning task that aims to transfer knowledge from a labeled source domain to an unlabeled target domain, addressing the costly challenge of large-scale data annotation in deep learning. Among mainstream UDA methods, some methods design and minimize specific metrics, such as maximum mean discrepancy\cite{gretton2006kernel}, to align the distributions of the source and target domains in a shared feature space; others acquire domain-invariant representations via adversarial learning \cite{jingWDAN,DomainPromptTuning}. 
Additionally, {KTransGAN \cite{azzam2020ktransgan} reduces domain discrepancy through collaborative learning between a classifier and a class-conditional generator, enabling the classifier to generalize effectively to the target domain.} 
Other approaches involve building cross-domain clustering centers or category prototypes\cite{CDCL,tang2020unsupervised} for class alignment. However, UDA relies on impractical assumptions, such as the simultaneous availability of the source and target domains. When labeled source data is unavailable, these methods become inapplicable.

\subsection{Source-free Domain Adaptation}
Source-free domain adaptation (SFDA) is introduced to address the challenge of domain adaptation when source data is unavailable. Research on SFDA can be categorized into two main branches. The first branch\cite{kundu2020universal,kundu2020towards} takes a straightforward approach by generating a proxy source domain, transforming the SFDA problem into a UDA setting. The quality of the generated domain is crucial, but measuring the similarity between the proxy and the real source domain is challenging. The second branch focuses on pseudo-label-based self-training\cite{liang2020we,qu2022bmd}, where the source pre-trained model is fine-tuned using pseudo-labeled target samples. However, due to domain shift, pseudo labels assigned by the pre-trained model can be noisy. Numerous strategies\cite{liang2020we,chen2022self,ahmed2022cleaning} have been proposed to refine pseudo labels, yet noise remains an issue even after refinement.
TPDS \cite{tang2024source} attribute the limitations of existing methods to the lack of effective error-mitigation mechanisms when mining error-prone auxiliary information. 
Moreover, some methods \cite{Tian2024,zhang2022divide,liang2021source,10438926} filter out a subset of clean samples and use these reliable samples to guide the learning of the rest through semi-supervised learning or local consistency. 
{Similarly, an early intra-domain adaptation method \cite{pan2020unsupervised} separates target data into an easy and hard subsets using an entropy-based ranking function, followed by adversarial learning \cite{tzeng2017adversarial} to reduce the intra-domain gap. However, more recent intra-domain adaptation methods \cite{lu2024mlnet,yang2024learningwA} do not necessarily partition the target domain; instead, they use invariant feature learning with self-adaptive neighbor selection \cite{lu2024mlnet} or multi-view clustering consistency constraints \cite{yang2024learningwA} to alleviate intra-domain shifts.} 
Nevertheless, the majority of samples, aside from the clean ones or easy split, remain underutilized. To address this, our proposed method not only divides the target domain by identifying clean samples but also applies tailored learning strategies to both clean and remaining samples. Inspired by the observation that deep networks tend to learn clean samples faster than noisy ones\cite{zhang2017understanding}, we make our learning approach progressive, starting with the easy-to-adapt subdomain and advancing to the more challenging one. 

The most related works are BETA\cite{yang2023divide} and SEAL\cite{xia2024separation}, both of which are domain-division based black-box domain adaptation frameworks. These methods use cross-entropy loss between source-predicted labels and target model outputs for domain division, while our proposed DPL selects clean samples based on the confidence and uncertainty of predictions from the source pre-trained model. Additionally, the subsequent adaptation strategies differ. DPL consists of two alternating progressive learning stages, employing techniques such as uncertainty-aware self-training, contrastive learning, and soft-voting. In contrast, BETA relies on two-network co-training and the MixMatch semi-supervised learning approach\cite{Li2020DivideMix}. Although both DPL and SEAL use graph contrastive learning for subdomain alignment, their implementations are distinct. These differences highlight uniqueness of DPL and contribute to its effectiveness.

{Curriculum domain adaptation \cite{zhang2017curriculum, sakaridis2018model} is a related but distinct paradigm from our method. Instead of focusing on easy samples first, it selects easy and useful tasks to form a curriculum that helps the model learn necessary properties for adapting to the hard target domain \cite{zhang2017curriculum}. This approach often requires synthetic data and intermediate tasks. For example, in the context of challenging foggy scene understanding, CMAda \cite{sakaridis2018model} adapts a semantic segmentation model by first training on non-foggy images, then on synthetic light foggy images, and finally on real heavy foggy images.}   
\subsection{Contrastive Learning}
Contrastive learning\cite{oord2018representation}, a popular self-supervised approach, maximizes the similarity between similar data points while minimizing it between dissimilar ones. This enables deep models to learn robust feature representations without the need for data annotation.  
It has been used to to capture structural information in domain adaptation. By constructing a contrastive loss, CSCA\cite{dai2020contrastively} encourages features of the same class from both the source and target domains to be close in the latent space, while pushing apart features from different classes.
For SFDA, HCL\cite{huang2021model} employs historical model snapshots from the target domain for contrastive learning to obtain discriminative feature representations.
CREL\cite{zhang2023class} utilizes contrastive learning by comparing samples from different classes to enforce separation in the feature space.
SEAL\cite{xia2024separation} minimizes the distance between samples of the same class across subdomains using graph contrastive learning.
In DPL, we design three distinct types of contrastive learning: one to align the easy-to-adapt and hard-to-adapt subdomains, another to construct consistency loss for the easy-to-adapt, and a third to extract topological information from the hard-to-adapt. Subsection IV.F Ablation Study demonstrates the effectiveness of these approaches. 
\begin{figure*}[t!]
\centering
\includegraphics[width=1\textwidth]{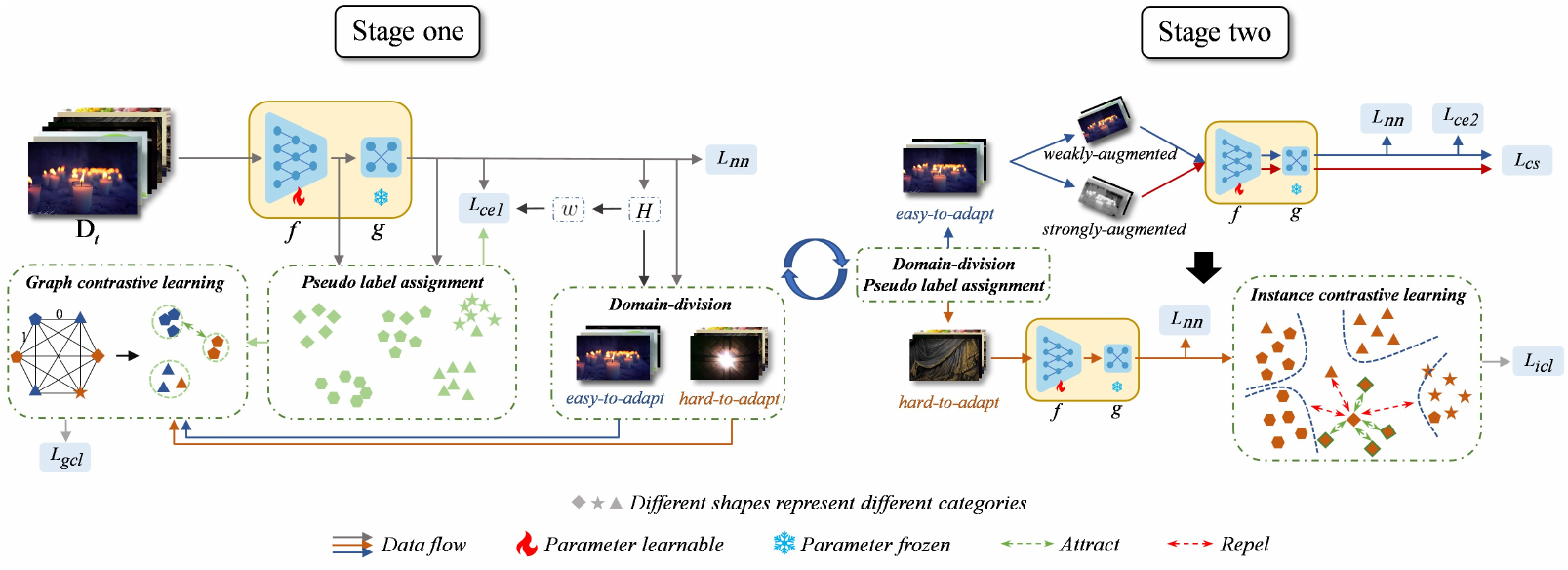}
\caption{{The overview of our proposed DPL.}
Its training process comprises of two alternating stages. Each stage begins with domain-division operation, followed by pseudo label assignment. Stage one incorporates uncertainty-aware self-training and subdomain alignment through graph contrastive learning. To leverage the intrinsic properties of the target data, stage two then applies tailored learning strategies to each subdomain, starting with consistency learning on the easy-to-adapt samples and progressing to mining local structural information for the more challenging ones. 
}
\label{fig:f2}
\end{figure*}

\section{Methodology }
\subsection{Preliminaries}
For SFDA, we are given a source pre-trained model consisting of a learnable feature extractor $f$ and a frozen classifier $g$. The unlabeled target domain is denoted as $\mathcal{D}_t=\{x_i\}_{i=1}^{N_t}$, where ${N_t}$ is the number of target samples. The goal is to adapt the model to the $\mathcal{D}_t$. In our setting, the source and target domains are assume to share the same class set containing $C$ categories. We represent the feature vector of a target sample $x_i$ as $q_t=f(x_i)$, and the softmax output of the model as $p_t=\delta(g(f(x_i)))$, where $\delta$ is the softmax function. The overall framework of our proposed method is illustrated in the Fig.\ref{fig:f2}. The model adaptation process consists of two alternating stages, which repeat until the model converges. At the start of each stage, we construct two memory banks to store the feature vectors and softmax outputs of all target samples, denoted as $F=\{q_1,q_2,…,q_{N_t}\}$ and $P=\{p_1,p_2,…,p_{N_t}\}$, respectively. 
Before detailing the learning strategies in each stage, we first introduce the domain division and pseudo label assignment that precede each stage.

\subsection{Domain-division Strategy}
Previous studies\cite{arpit2017closer,zhang2017understanding} have shown that deep models tend to prioritize learning simple patterns and fit clean samples more quickly than noisy ones.
We find that the source pre-trained model can still achieve good classification performance for a part of the target samples in the presence of domain shift. 
Based on this observation, we design a mechanism that divides the target domain according to the performance of the model. Specifically, samples classified correctly with high confidence and low uncertainty  are assigned to the easy-to-adapt subdomain, denoted as $E$. The remaining samples are put to the hard-to-adapt one, denoted as ${H}$. Some categories might be very easy to classify. To prevent the $E$ from being dominated by several easy categories, we conduct the division in a class-wise way, consisting of two steps. 

In the first step, we sort all target samples according to their predicted confidence scores on category $c$, $p_i^c$, that is the possibilities on $c$ the model predict. Then we select the top $M$ samples as candidates for class $c$ in the easy-to-adapt subdomain, where $M=N_t/C$. The set containing the candidates of $c$ can be denoted as follows,
\begin{equation}
    E'_c = \{ \, x_i \, | \, x_i \in topM(p^c_i)\},
    \label{set of candidate easy samples of class c}
\end{equation}
where $topM(p^c_i)$ denotes the sample set obtained via operation of sorting and selection based on $p^c_i$. 

In the second step, based on $E'_c$, we select the $N$ samples with the lowest information entropy of $p_i$, which quantifies the uncertainty of the predictions. Here, $N = \gamma \cdot M$, where $\gamma$ is a hyperparameter assigned different values depending on the dataset. The entropy of $p_i$, $H(p_i)$,is computed as follows, 
\begin{equation}
\begin{gathered}
    H(p_i) = -\,\sum_{c=1}^{C} p^c_i \, log \, p^c_i\,,\,1\le i \le N_t,\\
\end{gathered}
\label{entropy computation}
\end{equation}
Then, the easy-to-adapt set corresponding to category $c$ is defined as follows,
\begin{equation}
\begin{gathered}
    E_c = \{ \, x_i \, | \, x_i \in E'_c \ \& \ x_i \in {bottomN(H(p_i))}
    \},  
\end{gathered}
\end{equation}
where {$bottomN(H(p_i))$} denotes the set in which $N$ samples have the lowest entropy.

Finally, we can obtain the easy-to-adapt subdomain of the target domain as follows, 
\begin{equation}
\begin{gathered}
    E = E_1 \cup E_2 \cup … \cup E_C,\\
\end{gathered}
\end{equation}
then the rest is the hard-to-adapt one:
\begin{equation}
\begin{gathered}
    H = \mathcal{D}_t - E.
    \label{Hard-to-Adapt}
\end{gathered}
\end{equation}
Our proposed domain division relies on the model predictions, and as the model continually adapts to the target domain, the quality of these predictions improves. Therefore, we perform domain division at the beginning of each stage.

\subsection{Pseudo Label Assignment}
Pseudo labeling is a promising technique that is widely employed to tackle the issue of training with unlabeled data. We assign pseudo labels to target samples with a strategy similar to SHOT\cite{liang2020we}. 
The first step is to generate cluster centers for each category using the samples of easy-to-adapt subdomain:

\begin{equation}
    \mu^e_c=\frac{\sum_{i=1}^{n^e_c} p^c_i\,q_i}{\sum_{i=1}^{n^e_c} p^c_i},\,
    1 \le c \le C,
    \label{cluster_center}
\end{equation}
where $n^e_c$ is the number of samples classified as $c$ by the model in easy-to-adapt subdomain, $p_i$ and $q_i$ are the softmax output and {normalized low-dimensional} feature vector of sample $x_i$, respectively.
The cluster centers computed in this way characterize the category distribution of the target domain. Subsequently, We calculate the cosine distance between each sample and all cluster centers, assigning the pseudo label of the category with the smallest distance to each sample：
\begin{equation}
    \hat{y}_i = arg\, \mathop{min}\limits_c D(q_i, \mu^e_c),
    \label{pseudo_argmin}
\end{equation}
where $D(\cdot,\cdot)$ is a function of the cosine distance, $\hat{y}_i$ is the pseudo label of the sample $x_i$. 
We observe the above pseudo label assignment approach is effective in the early phase of training. However, it is also influenced by global noise. As training progresses, some hard-to-adapt samples can cause error accumulation, leading to limited improvement in accuracy of pseudo labels. Actually, the pseudo assignment introduced above is only adopted in the first epoch of training.  
Afterwards, we turn to utilize the cluster assumption that samples located closely with each other are likely to possess the same label. Inspired by works \cite{mitchell2001soft,zhang2023class}, we switch to a neighborhood based soft voting to obtain pseudo labels. Specifically, the softmax output of $T$ neighboring samples are used to synthesize the prediction for sample $x_i$:
\begin{equation}
\begin{gathered}
    \bar{p}_i = \frac{1}{T} \sum_{j=1}^T\,p_j,\\
\end{gathered}
\label{mean_neigb_predict}
\end{equation}

\begin{equation}
\begin{gathered}
    \hat{y}_i = arg \mathop{max}\limits_c \bar{p}^c_i.
\end{gathered}
\label{argmax_c_predict}
\end{equation}

Similar to domain division, pseudo label assignment is performed at the start of each stage, and both processes are carried out without gradient updates.
\subsection{Stage One}
In stage one, the model is first optimized through uncertainty-aware self-training with pseudo labels, then the corresponding categories of two subdomains are aligned through graph contrastive learning.  

\subsubsection{Uncertainty-aware Self-training}
The model can be self-trained using the obtained pseudo labels. However, since pseudo labels inevitably contain noise, we need to minimize its impact during self-training when they may be inaccurate. A pseudo label $\hat{y}_i$ is likely to be unreliable if the uncertainty of model prediction $p_i$ is high. To address this, we compute a weight $w_i$ for each sample $x_i$ based on the entropy of $p_i$, which quantifies the uncertainty:
\begin{equation}
\begin{gathered}
    w_i = exp(-\frac{H(p_i)}{logC}),
\end{gathered}
\label{weight}
\end{equation} 
where the entropy $H(p_i)$ can be computed following the equation (\ref{entropy computation}). 
In Equation (\ref{weight}), the entropy $H(p_i)$ is first normalized and then inversely mapped to $w_i$. This weight is integrated into the uncertainty-aware self-training loss as follows:
\begin{equation}
    L_{ce1} = -\,\frac{1}{N_t}\sum^{N_t}_{i=1} w_i \sum^{C}_{c=1}\hat{y}^c_i log p^c_i,
\end{equation} 
where $w_i$ reduces the influence of samples with unreliable pseudo labels.

To prevent model collapse and improve the diversity of the model prediction, we adopt a nuclear norm maximization loss\cite{cui2020towards}. It is calculated using a mini-batch of samples $\left\{x_i\right\}_{i=1}^{N_b}$ as follows,
\begin{equation}
    L_{nn} = - \sqrt{\sum^{N_b}_{i=1}\sum^{C}_{j=1}\mid X_{ij}\mid^2},
    \label{nn_loss}
\end{equation}
where the matrix $X$ denotes the softmax output of a mini-batch of samples.

\subsubsection{Graph Contrastive Learning for Subdomain Alignment}
The uncertainty-aware self-learning described above primarily utilizes easy-to-adapt samples, leaving hard-to-adapt samples underexploited and their predictions unimproved. To better leverage and adapt the hard-to-adapt subdomain, we align the two subdomains in feature space, where samples naturally form clusters. According to the cluster assumption, samples of the same category should be located closely together. However, since the pseudo labels of hard-to-adapt samples are often unreliable, naively clustering them with the same pseudo labels through supervised contrastive learning\cite{khosla2020supervised,jing2024NENO} can lead to compounded prediction errors.

{Inspired by SEAL \cite{xia2024separation}}, we propose to align subdomains using graph contrastive learning based on a label graph and a feature graph. Specifically, for a mini-batch of target samples $\left\{x_i\right\}_{i=1}^{N_b}$, we first construct a label graph $R$, where each node represents a sample $x_i$. The weight $R_{ij}$ of the edge in the label graph is defined as follows:
\begin{equation}
\begin{gathered}
    R_{ij} = \left\{
        \begin{array}{cc}
             1,&  ({x_i\in E\, \vee\, x_j\in E}) \wedge \hat{y}_i=\hat{y}_j,\\
             0,&  {otherwise}.
        \end{array} 
             \right.
\label{LabelGraph}             
\end{gathered}
\end{equation}
This construction connects samples of the same potential classes within the easy-to-adapt subdomain and across both subdomains, serving as the training target for the feature graph to facilitate subdomain alignment via feature similarity. Additionally, we fill in ones on the diagonal of $R$ to encourage self-alignment for all samples.  
Unlike the label graph $R$ {defined in equation (\ref{LabelGraph})}, the weight $A_{ij}$ of the edge in the feature graph $A$ represents the similarity of features between samples $x_i$ and $x_j$:
\begin{equation}
\begin{gathered}
    A_{ij} = q_i \cdot q_j.
\end{gathered}
\end{equation}
The diagonal of $A$ is also filled with 1. 
Finally, the goal of graph contrastive learning is defined as the cross-entropy loss between the label graph and the feature graph:
\begin{equation}
    L_{gcl} = - \frac{1}{N_b}\sum^{N_b}_{i=1}\sum^{N_b}_{j=1} R_{ij} log A_{ij},
\end{equation}	
where both graphs are row-wise normalized.

\subsection{Stage Two} 
In stage two, we further exploit the intrinsic properties of the target data through applying tailored learning strategies to each subdomain, beginning with the easy-to-adapt then moving to the hard-to-adapt one. 
For easy-to-adapt samples, pseudo labels are reliable, allowing for standard self-training. Additionally, consistency learning based on different augmentations of a sample is employed to enhance feature representation. 
However, pseudo labels for hard-to-adapt samples are often unreliable. Thus, we resort to unsupervised methods to leverage neighborhood information, avoiding reliance on pseudo labels.
\subsubsection{Self-training and Consistency Loss on Easy Samples}
As pseudo labels of easy-to-adapt samples are reliable, standard self-training can be conducted as follows, 
\begin{equation}
    L_{ce2} = -\,\frac{1}{|E|}\sum^{|E|}_{i=1}\sum^{C}_{c=1}\hat{y}^c_i log p^c_i
    \label{standard_selftraning_loss}
\end{equation}
Consequently, the model memory of these samples is strengthened.
Furthermore, the consistency learning of different augmentations is utilized to improve the quality of feature representation. $p_i$ mentioned above refers to the softmax output of the sample $x_i$ using a weak augmentation, while $p^{\prime}_i$ represents the softmax output of the sample using a strong augmentation. We define the consistency loss between different augmentations as follows,
\begin{equation}
    L_{cs} = -\,\frac{1}{|E|}\sum^{|E|}_{i=1}\sum^{C}_{c=1} p^c_i log {p^{\prime}}^c_i.
    \label{loss_cs}
\end{equation}

\subsubsection{Instance Contrastive Learning on Hard Samples}
The unreliable pseudo labels for hard-to-adapt samples necessitate an approach that does not depend on these labels. 
Unlike the graph contrastive learning in phase one, which processes data solely within a mini-batch, we take into account the entire hard-to-adapt subdomain and propose an instance contrastive learning utilizing local structural information. 
Contrastive learning usually takes sample features as input to learn discriminatory representation, while disregards the domain information contained in the classifier. Previous work\cite{li2024ijcai} finds that calculating the contrastive loss in the output space can alleviate this problem. Hence, we use the softmax outputs $P$ of samples as input. 
Despite the unreliable predictions for hard samples, their features still retain class semantic information. Therefore, identifying neighbors based on these features and using this information to regularize their outputs remains effective. 
For a hard-to-adapt sample $x^h_i$ whose softmax output is $p^h_i$, we first obtain its $T$ nearest neighbors throughout the entire hard-to-adapt subdomain based on cosine distance, denoted as $\mathcal{N}(x^h_i)$. Then we take the mean of softmax output of these neighbors as the positive example for $p^h_i$:  
\begin{equation}
    \begin{gathered}
        p^{pos}_i = \frac{1}{T} \sum_{k=1}^T\,p_k, x_k\in \mathcal{N}(x^h_i),\\
    \end{gathered}
    \label{PositiveExample}
\end{equation}
and the softmax outputs of the rest samples as negative examples for $p^h_i$:
\begin{equation}
    \begin{gathered}
        p^{neg}_{ij} \in \{p_j\, |\, x_j \notin \mathcal{N}(x^h_i) \wedge x_j \in H\}.
    \end{gathered}
\end{equation}

Subsequently, we maximize the similarity with the positive example while minimizing the similarity with the negative examples. The instance contrastive loss is defined as follows, 
\begin{equation}
    \begin{gathered}
        L_{icl} = -\,\frac{1}{|H|}\sum^{|H|}_{i=1}log\frac{s^{pos}_i}{s^{pos}_i + s^{neg}_i},
    \end{gathered}
    \label{loss_igl}
\end{equation}
where $s^{pos}_i$ and $s^{neg}_i$ measure the similarity between $p^h_i$ and the positive and negative examples, respectively. These similarities are computed using the inner product as follows:
\begin{equation}
    \begin{gathered}
        s^{pos}_i = exp(p_i^h \cdot p^{pos}_i / \tau),\\
    \end{gathered}
\end{equation}
\begin{equation}
    \begin{gathered}
        s^{neg}_i = \sum^{|H-\mathcal{N}(x^h_i)|}_{j=1}exp(p_i^h \cdot p^{neg}_{ij} / \tau),
    \end{gathered}
\end{equation} 
where $\tau$ is the temperature parameter, set to 0.07 throughout the experiment. 

Additionally, the nuclear norm maximization loss, defined in equation (\ref{nn_loss}), is also used in stage two.

\subsection{Overall Loss of Each Training Stage}
The proposed stages one and two are trained alternately. The training loss for stage one is defined as follows,
\begin{equation}
    L_{s1} = L_{ce1} + \lambda\cdot L_{gcl} + \alpha \cdot L_{nn}.
    \label{Stage1-loss}
\end{equation}
The training loss for stage two is defined as:
\begin{equation}
    L_{s2} = L_{ce2} + L_{cs} + L_{icl} + \beta \cdot L_{nn}
    \label{Stage2-loss}
\end{equation}
where $\lambda$, $\alpha$, and $\beta$ are coefficients used to balance the respective losses. The overall training procedure is outlined in Algorithm \ref{alg1}.

\begin{algorithm}[t!]
    \caption{Training procedure of DPL}
    \renewcommand{\algorithmicrequire}{\textbf{Input:}}
    \renewcommand{\algorithmicensure}{\textbf{Output:}}
    \label{alg1}
    \begin{algorithmic}[1]
        \REQUIRE Unlabeled target data $\mathcal{D}_t=\{x_i\}_{i=1}^{N_t}$; 
        the source pre-trained model consisting of a learnable feature extractor $f$ and a frozen classifier $g$.\\
        \ENSURE Target model \{$f_t$, $g_t$\}

        \renewcommand{\algorithmicensure}{\textbf{Parameters:}}

        \ENSURE training epoch $e$;\\ target data iterations (mini-batch number) $N_i$ per epoch;\\ easy-to-adapt data iterations $N_e$ per epoch;\\ hard-to-adapt data iterations $N_h$ per epoch.
        
        \STATE Initialization $f_t$ = $f$, $g_t$ = $g$, ${P}$, ${F}$. 
        \FOR{epoch-num = 1 \textbf{to} $e$}
            \STATE Update memory banks ${P}$, ${F}$.
            
            \STATE Divide target data into ${E}$,${H}$ via Eq.(\ref{set of candidate easy samples of class c})-Eq.(\ref{Hard-to-Adapt}).
            
            \IF{epoch-num = 1}
                \STATE Obtain pseudo labels via Eq.(\ref{cluster_center})-Eq.(\ref{pseudo_argmin}).
            \ELSE
                \STATE Obtain pseudo labels via Eq.(\ref{mean_neigb_predict})-Eq.(\ref{argmax_c_predict}).
            \ENDIF
            
            \FOR{iter-num = 1 \textbf{to} $N_i$}
            
            \STATE Update model $f_t$ via minimizing Eq.(\ref{Stage1-loss}).
            
            \ENDFOR
            
            \STATE Update memory banks ${P}$, ${F}$.
            
            \STATE Divide target data into ${E}$,${H}$ via Eq.(\ref{set of candidate easy samples of class c})-Eq.(\ref{Hard-to-Adapt}).
            
            \IF{epoch-num = 1}
                \STATE Obtain pseudo labels via Eq.(\ref{cluster_center})-Eq.(\ref{pseudo_argmin}).
            \ELSE
                \STATE Obtain pseudo labels via Eq.(\ref{mean_neigb_predict})-Eq.(\ref{argmax_c_predict}).
            \ENDIF
            
            \FOR{iter-num = 1 \textbf{to} $N_e$}
            
            \STATE Update model $f_t$ via minimizing Eq.(\ref{standard_selftraning_loss}), Eq.(\ref{loss_cs}), and Eq.(\ref{nn_loss}).
            
            \ENDFOR
            
            \FOR{iter-num = 1 \textbf{to} $N_h$}
            
            \STATE Update model $f_t$ via minimizing Eq.(\ref{loss_igl}) and Eq.(\ref{nn_loss}).
            
            \ENDFOR
        \ENDFOR
    \end{algorithmic}
\end{algorithm}

\section{Experiments }

\subsection{Datasets}
We assessed the effectiveness of our method on three widely used domain adaptation benchmarks:

\textbf{Office-31}\cite{saenko2010adapting} is a small-scale dataset commonly used for domain adaptation. It comprises three different domains: Amazon (A), DSLR (D), and Webcam (W), each containing 31 categories and a total of 4,652 images. 

\textbf{Office-Home}\cite{venkateswara2017deep} is a more challenging medium-sized dataset that includes 15,000 images across 65 categories from work and home environments. It consists of four distinct domains: Artistic (Ar), Clipart (Cl), Product (Pr), and Real World (Rw).

\textbf{VisDA}\cite{peng2017visda} comprises virtual and real-world images, including a total of 12 categories. The source domain contains 152,000 synthetic images, while the target domain includes 55,000 real-world images.

\begin{table}[!t]\renewcommand\arraystretch{1.3}
\caption{
Results (\%) on Office-31. 
``SF'' is short for ``source-free'', ``Source-only'' denotes the source-pretrained model, and ``Avg.'' means average accuracy. The best results are in bold.   
}
\centering
\scalebox{0.8}{
\begin{tabular}{l|c|ccccccc} 
\hline
 Method& SF& A$\rightarrow$D& A$\rightarrow$W& D$\rightarrow$A& D$\rightarrow$W& W$\rightarrow$A& W$\rightarrow$D& Avg.\\ 
\hline
 Source-only& - & 48.2& 67.0& 74.1& 49.2& 62.0& 62.7&60.4\\
 \hline
 CDAN\cite{long2018conditional}& \usym{2717} & 92.9 & 94.1 & 71.0 & 98.6 & 69.3 & 100.0 &87.7 
\\
 SAFN\cite{xu2019larger}& \usym{2717} & 95.6 & 95.4 & 72.6 & 98.6 & 73.9 & 100.0 &89.4 
\\
 IA\cite{jiang2020implicit}& \usym{2717} & 92.1 & 90.3 & 75.3 & 98.7 & 74.9 & 99.8 &88.8 
\\
 MSGD\cite{xia2022maximum}& \usym{2717} & 91.2 & 90.8 & 72.2 & 98.7 & 71.4 & 100.0 &87.4 
\\
 AML\cite{zhou2023adaptive}& \usym{2717} & 92.4 & 94.2 & 75.5 & 98.9 & 75.9 & 100.0 &89.5 
\\
 MLSL\cite{zhu2024multiview}& \usym{2717} & 95.8 & 94.5 & 76.5 & 99.0 & 78.3 & 99.8 &90.7 
\\
\hline
 SHOT\cite{liang2020we}& \usym{1F5F8} & 94.0 & 90.1 & 74.7 & 98.4 & 74.3 & 99.9 &88.6 
\\
 SHOT++\cite{liang2021source}& \usym{1F5F8} & 94.3 & 90.4 & 76.2 & 98.7 & 75.8 & 99.9 &89.2 
\\
 NRC\cite{yang2021exploiting}& \usym{1F5F8} & 96.0 & 90.8 & 75.3 & 99.0 & 75.0 & \textbf{100.0} &89.4 
\\
 DIPE\cite{wang2022exploring}& \usym{1F5F8} & 96.6 & 93.1 & 75.5 & 98.4 & 77.2 & 99.6 &90.1 
\\
 AaD\cite{yang2022attracting}& \usym{1F5F8} & 96.4 & 92.1 & 75.0 & \textbf{99.1} & 76.5 & \textbf{100.0} &89.9 
\\
 CoNMix\cite{kumar2023conmix}& \usym{1F5F8} & 88.8 & 94.0 & \textbf{77.3} & 98.1 & 75.2 & \textbf{100.0} &88.9 
\\
 UTR\cite{pei2023uncertainty}& \usym{1F5F8} & 95.0 & 93.5 & 76.3 & \textbf{99.1} & \textbf{78.4} & 99.9 &90.3 
\\
 TPDS\cite{tang2024source}& \usym{1F5F8} & \textbf{97.1} & 94.5 & 75.7 & 98.7 & 75.5 & 99.8 &90.2 
\\
 UMNC\cite{xiao2024unified}& \usym{1F5F8} & 96.2 & \textbf{94.7} & 76.1 & 98.9 & 76.3 & \textbf{100.0} &90.4 
\\
 DPL (Ours)& \usym{1F5F8} & 97.0 & 93.5 & 77.0 & 98.6 & 78.0 & \textbf{100.0} & \textbf{90.7}
\\
\hline\end{tabular}
}
\label{tab:t1}
\end{table}

\begin{table*}[htb]\renewcommand\arraystretch{1.3}
\caption{Results (\%) on Office-Home. 
``SF'' is short for ``source-free'', ``Source-only'' denotes the source-pretrained model, and ``Avg.'' represents average accuracy across all the tasks. The best results are highlighted in bold.}
\centering
\scalebox{0.9}{
\begin{tabular}{l|c|cccccccccccc|c} 
\hline
 Method&  \
 SF& \
 Ar$\rightarrow$Cl& Ar$\rightarrow$Pr& Ar$\rightarrow$Rw& Cl$\rightarrow$Ar& Cl$\rightarrow$Pr& Cl$\rightarrow$Rw& Pr$\rightarrow$Ar& Pr$\rightarrow$Cl& Pr$\rightarrow$Rw& Rw$\rightarrow$Ar& Rw$\rightarrow$Cl& Rw$\rightarrow$Pr& Avg.\\ 
\hline
 Source-only& - & 48.2& 67.0& 74.1& 49.2& 62.0& 62.7& 51.8& 42.6& 72.6& 64.6& 51.5& 78.7&60.4\\
 \hline
 CDAN\cite{long2018conditional}& \usym{2717} & 50.7 & 70.6 & 76.0 & 57.6 & 70.0 & 70.0 & 57.4 & 50.9 & 77.3 & 70.9 & 56.7 & 81.6 &65.8 
\\
 SAFN\cite{xu2019larger}& \usym{2717} & 52.0 & 71.7 & 76.3 & 64.2 & 69.9 & 71.9 & 63.7 & 51.4 & 77.1 & 70.9 & 57.1 & 81.5 &67.3 
\\
 IA\cite{jiang2020implicit}& \usym{2717} & 56.0 & 77.9 & 79.2 & 64.4 & 73.1 & 74.4 & 64.2 & 54.2 & 79.9 & 71.2 & 58.1 & 83.1 &69.5 
\\
 MSGD\cite{xia2022maximum}& \usym{2717} & 58.7 & 76.9 & 78.9 & 70.1 & 76.2 & 76.6 & 69.0 & 57.2 & 82.3 & 74.9 & 62.7 & 84.5 &72.4 
\\
 AML\cite{zhou2023adaptive}& \usym{2717} & 58.9 & 77.2 & 81.7 & 69.6 & 77.9 & 78.6 & 66.6 & 57.9 & 82.3 & 74.7 & 62.5 & 84.5 &72.7 
\\
 MLSL\cite{zhu2024multiview}& \usym{2717} & 56.2 & 80.5 & 81.6 & 67.4 & 78.3 & 80.6 & 66.3 & 54.1 & 82.9 & 72.6 & 57.1 & 86.6 &72.0 
\\
\hline
 SHOT\cite{liang2020we}& \usym{1F5F8} & 56.6 & 78.0 & 80.6 & 68.4 & 78.1 & 79.4 & 68.0 & 54.3 & 82.2 & 74.3 & 58.7 & 84.5 &71.8 
\\
 SHOT++\cite{liang2021source}& \usym{1F5F8} & 57.9 & 79.7 & 82.5 & 68.5 & 79.6 & 79.3 & 68.5 & 57.0 & 83.0 & 73.7 & 60.7 & 84.9 &73.0 
\\
 NRC\cite{yang2021exploiting}& \usym{1F5F8} & 57.7 & 80.3 & 82.0 & 68.1 & 79.8 & 78.6 & 65.3 & 56.4 & 83.0 & 71.0 & 58.6 & 85.6 &72.2 
\\
 DIPE\cite{wang2022exploring}& \usym{1F5F8} & 56.5 & 79.2 & 80.7 & 70.1 & 79.8 & 78.8 & 67.9 & 55.1 & \textbf{83.5} & 74.1 & 59.3 & 84.8 &72.5 
\\
 AaD\cite{yang2022attracting}& \usym{1F5F8} & 59.3 & 79.3 & 82.1 & 68.9 & 79.8 & 79.5 & 67.2 & \textbf{57.4} & 83.1 & 72.1 & 58.5 & 85.4 &72.7 
\\
 CoNMix\cite{kumar2023conmix}& \usym{1F5F8} & 57.6 & 77.2 & 82.2 & 68.4 & 78.8 & 78.3 & 67.1 & 54.7 & 81.5 & 74.0 & 60.2 & 85.3 &72.1 
\\
 UTR\cite{pei2023uncertainty}& \usym{1F5F8} & 59.8 & \textbf{81.2} & \textbf{83.2} & 67.2 & 79.2 & 80.1 & 68.4 & 56.4 & 83.0 & 73.7 & 61.2 & 85.9 &73.2 
\\
 TPDS\cite{tang2024source}& \usym{1F5F8} & 59.3 & 80.3 & 82.1 & \textbf{70.6} & 79.4 & \textbf{80.9} & 69.8 & 56.8 & 82.1 & 74.5 & 61.2 & 85.3 &73.5 
\\
 UMNC\cite{xiao2024unified}& \usym{1F5F8} & 59.2 & 80.8 & 81.3 & 70.0 & \textbf{80.9} & 79.5 & 68.6 & 57.2 & 82.7 & 74.6 & 60.2 & 85.7 &73.4 
\\
 DPL (Ours)& \usym{1F5F8} & \textbf{60.9} & 80.8 & 82.3 & 69.9 & 79.5 & 80.0 & \textbf{70.0} & 57.0 & 82.7 & \textbf{75.7} & \textbf{63.0} & \textbf{86.0} & \textbf{74.0}
\\
\hline\end{tabular}
}
\label{tab:t2}
\end{table*}

\begin{table*}[htb]\renewcommand\arraystretch{1.3}
\caption{Results (\%) on VisDA. 
``SF'' is short for ``source-free'', ``Source-only'' denotes the source-pretrained model, and ``Avg.'' represents average accuracy across all the tasks. The best results are highlighted in bold.}
\centering
\scalebox{1.1}{
\begin{tabular}{l|c|cccccccccccc|c} 
\hline
 Method&  \
 SF& plane&  bcycl&  bus&  car&  horse&  knife&  mcycl&  person&  plant&  sktbrd&  train&  truck&  Avg.\\ 
\hline
 Source-only& - & 63.7& 14.9& 53.0& 69.9& 70.5& 5.5& 79.4& 31.0& 77.5& 37.1& 83.4& 5.8&49.3\\
 \hline
 CDAN\cite{long2018conditional}& \usym{2717} & 85.2 & 66.9 & 83.0 & 50.8 & 84.2 & 74.9 & 88.1 & 74.5 & 83.4 & 76.0 & 81.9 & 38.0 &73.9 
\\
 SAFN\cite{xu2019larger}& \usym{2717} & 93.6 & 61.3 & 84.1 & 70.6 & 94.1 & 79.0 & 91.8 & 79.6 & 89.9 & 55.6 & 89.0 & 24.4 &76.1 
\\
 IA\cite{jiang2020implicit}& \usym{2717} & -& -& -& -& -& -& -& -& -& -& -& -&75.8 
\\
 MSGD\cite{xia2022maximum}& \usym{2717} & 97.5 & 83.4 & 84.4 & 69.4 & 95.9 & 94.1 & 90.9 & 75.5 & 95.5 & 94.6 & 88.1 & 44.9 &84.6 
\\
 AML\cite{zhou2023adaptive}& \usym{2717} & 96.7 & 88.5 & 79.6 & 69.0 & 95.9 & 96.3 & 87.3 & 83.3 & 94.4 & 92.9 & 87.0 & 58.7 &85.8 
\\
\hline
 SHOT\cite{liang2020we}& \usym{1F5F8} & 94.3 & 88.5 & 80.1 & 57.3 & 93.1 & 94.9 & 80.7 & 80.3 & 91.5 & 89.1 & 86.3 & 58.2 &82.9 
\\
 SHOT++\cite{liang2021source}& \usym{1F5F8} & 97.7 & 88.4 & \textbf{90.2} & \textbf{86.3} & \textbf{97.9} & \textbf{98.6} & \textbf{92.9} & 84.1 & \textbf{97.1} & 92.2 & \textbf{93.6} & 28.8 &87.3 
\\
 GKD\cite{tang2021model}& \usym{1F5F8} & 95.3 & 87.6 & 81.7 & 58.1 & 93.9 & 94.0 & 80.0 & 80.0 & 91.2 & 91.0 & 86.9 & 56.1 &83.0 
\\
 NRC\cite{yang2021exploiting}& \usym{1F5F8} & 96.8 & 91.3 & 82.4 & 62.4 & 96.2 & 95.9 & 86.1 & 80.6 & 94.8 & 94.1 & 90.4 & 59.7 &85.9 
\\
 DIPE\cite{wang2022exploring}& \usym{1F5F8} & 95.2 & 87.6 & 78.8 & 55.9 & 93.9 & 95.0 & 84.1 & 81.7 & 92.1 & 88.9 & 85.4 & 58.0 &83.1 
\\
 AAA\cite{li2021divergence}& \usym{1F5F8} & 94.4 & 85.9 & 74.9 & 60.2 & 96.0 & 93.5 & 87.8 & 80.8 & 90.2 & 92.0 & 86.6 & 68.3 &84.2 
\\
 UTR\cite{pei2023uncertainty}& \usym{1F5F8} & 98.0 & 92.9 & 88.3 & 78.0 & 97.8 & 97.7 & 91.1 & 84.7 & 95.5 & 91.4 & 91.2 & 41.1 &87.3 
\\
 TPDS\cite{tang2024source}& \usym{1F5F8} & 97.6 & 91.5 & 89.7 & 83.4 & 97.5 & 96.3 & 92.2 & 82.4 & 96.0 & 94.1 & 90.9 & 40.4 &87.6 
\\
 UMNC\cite{xiao2024unified}& \usym{1F5F8} & 97.2 & 92.2 & 82.6 & 67.0 & 96.9 & 96.6 & 87.4 & 81.6 & 94.9 & 93.6 & 90.0 & 62.5 &86.9 
\\
 DPL (Ours)& \usym{1F5F8} & \textbf{98.4} & \textbf{93.7} & 84.2 & 58.9 & 97.0 & 97.5 & 85.7 & \textbf{85.2} & 95.1 & \textbf{96.0} & 92.8 & \textbf{69.1} & \textbf{87.8}
\\
\hline\end{tabular}
}
\label{tab:t3}
\end{table*}

\subsection{Implementation Details}
\textbf{Network architecture.}
All experiments are conducted using PyTorch 2.1.2 on RTX4090 24GB GPUs. To ensure fair comparison of relevant methods, we use the pre-trained ResNet50\cite{he2016deep} as the backbone network for Office-31 and Office-Home, and ResNet101\cite{he2016deep} as the backbone network for VisDA. Following SHOT\cite{liang2020we}, we replace the original fully connected(FC) layer with a bottleneck layer with 256 units. We utilize batch normalization following the bottleneck layer and apply WeightNorm\cite{salimans2016weight} to the classifier.

\textbf{Network hyperparameters.}
For model training on the source domain, we follow the settings of SHOT\cite{liang2020we} and initialize model parameters using pre-trained weights on ImageNet-1K\cite{deng2009imagenet}. For the bottleneck layer and classifier, we use a learning rate 10 times higher than the backbone network for training. We set the training epochs on Office-31, Office-Home and VisDA to 100, 50 and 10, respectively. 
For training on the target domain, we use stochastic gradient descent optimizer with momentum of 0.9 and weight decay of 1e-3. We set the learning rate to 1e-2 for Office-31 and Office-Home, 1e-3 for VisDA. We assign dataset specific values to hyperparameters. {For sample selection $\gamma$ used in the second step of domain division and $T$ whose value is the number of neighboring samples, we set $\gamma = 0.4/0.5/0.5$ and $T$= $5/4/5$ for Office-31, Office-Home, and VisDA, respectively.} To balance the loss terms in the equation (\ref{Stage1-loss}) and (\ref{Stage2-loss}), we use different nuclear-norm loss coefficients for different datasets. We set $\alpha=3,\beta=2$ for Office-31, {$\alpha=0,\beta=3$} for Office-Home, and $\alpha=0,\beta=0$ for VisDA. For $\lambda$, we use the value with the best performance, whilst $\lambda = 0.7/0.6/0.6$ for Office-31, Office-Home, and VisDA. The sensitivity analysis of hyperparameters is shown in the {Fig.\ref{fig:f4} and Fig.\ref{fig:ab}}.

\subsection{Compared Methods and Evaluation Metric}
In experiments, our proposed DPL is compared with state of the art UDA and SFDA methods. The UDA methods include CDAN\cite{long2018conditional}, SAFN\cite{xu2019larger}, IA\cite{jiang2020implicit}, MSGD\cite{xia2022maximum}, AML\cite{zhou2023adaptive}, MLSL\cite{zhu2024multiview}, MCC\cite{jin2020minimum}, and SUDA\cite{zhang2022spectral}. The SFDA methods include SHOT\cite{liang2020we}, SHOT++\cite{liang2021source}, NRC\cite{yang2021exploiting}, DIPE\cite{wang2022exploring}, AaD\cite{yang2022attracting}, CoNMiX\cite{kumar2023conmix}, UTR\cite{pei2023uncertainty}, TPDS\cite{tang2024source}, UMNC\cite{xiao2024unified}, GKD\cite{tang2021model}, and AAA\cite{li2021divergence}. 

Classification accuracy on the target domains of adaptation tasks serves as the evaluation metric.
    
\subsection{Experimental Results}
The experimental results on Office-31, Office-Home, and VisDA are presented in Table \ref{tab:t1}, Table \ref{tab:t2} and Table \ref{tab:t3}, respectively, with the best results highlighted in \textbf{bold}. 
The ``SF'' in the tables stands for source-free, while ``\usym{1F5F8}'' indicates that the corresponding method is an SFDA method, and ``\usym{2717}'' signifies that the method is a UDA method.
The baseline ``Source-only'' method directly use the source-pretrained model to predict categories of the target samples. 
All results for the compared methods in these tables are reported from the cited papers.

\textbf{Office-31}. As shown in Table \ref{tab:t1}, our approach outperforms most UDA methods requiring a source domain and achieves comparable performance to MLSL, the latest UDA method. While it may not attain the best performance on most tasks compared to other SFDA methods, our method achieve the highest average accuracy (Avg.) of 90.7\%, which is 0.3\% higher than UMNC and 0.5\% higher than TPDS, both of which are recent SFDA methods.

\textbf{Office-Home}. As presented in Table \ref{tab:t2}, our approach attains better Avg. than both UDA and SFDA compared methods. We obtain a Avg. of {74.0\%}, which is {0.5\%} higher than TPDS, {0.6\%} higher than UMNC, and {1.0\%} higher than SHOT++. Moreover, our method delivers the best results for half of the tasks. Notably, for the task of {Rw$\rightarrow$Cl, our accuracy is 1.8\% higher than TPDS, 2.8\% higher than UMNC, and 2.3\% higher than SHOT++.}

\textbf{VisDA}. As indicated in Table \ref{tab:t3}, our approach also demonstrates remarkable performance on a large-scale dataset like VisDA. With the exception of SHOT++, we achieve the highest accuracy in most categories, ultimately reaching an Avg. of 87.8\%, outperforming all compared UDA and SFDA methods. 
For the challenging category of trunk, our approach achieves the highest accuracy of 69.1\%. However, our method demonstrates mediocre performance in the car category, likely because samples in this category comprise approximately one-fifth of the entire target domain data. Excessive noisy samples may cause neighborhood clustering to incorrectly push car samples toward other categories. Notably, methods using neighborhood clustering, such as GKD, NRC, DIPE, and UMNC, show lower accuracy in this category compared to the source-only baseline.

As mentioned in the Introduction, TPDS emphasizes the importance of error-mitigation mechanisms. Meanwhile, our DPL outperforms TPDS in Avg. across all three datasets. This indicates that while clean samples are used to address the cross-domain distribution shift, our method also leverages the remaining noisy samples to learn the feature distribution, allowing the model to better exploit the target data. 

\begin{figure*}[t!]
\centering
\includegraphics[width=1\textwidth]{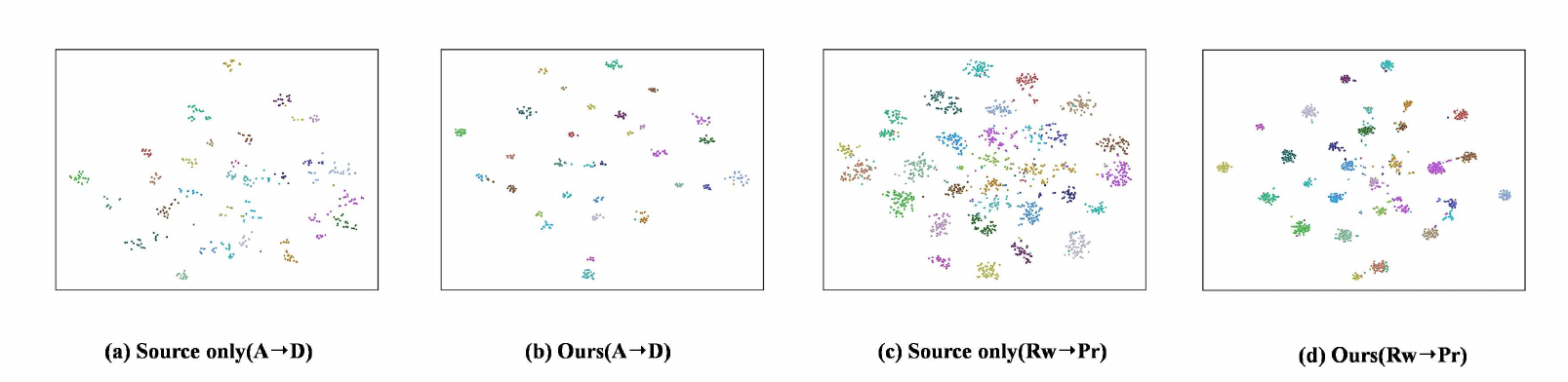}
\caption{t-SNE\cite{van2008visualizing} visualizations of the target domain features for A$\rightarrow$D on Office-31 and Rw$\rightarrow$Pr on Office-Home. Each point represents a sample, with different colors indicating distinct categories. The visualizations compare the feature distributions extracted by the source-only baseline and our proposed method. {For clarity, we select the first 30 classes from the total of 65 when visualizing the target domain of the Office-Home benchmark. (best viewed in color and magnified).}}
\label{fig:f3}
\end{figure*}

\begin{figure*}[t!]
\centering
\includegraphics[width=0.9\textwidth]{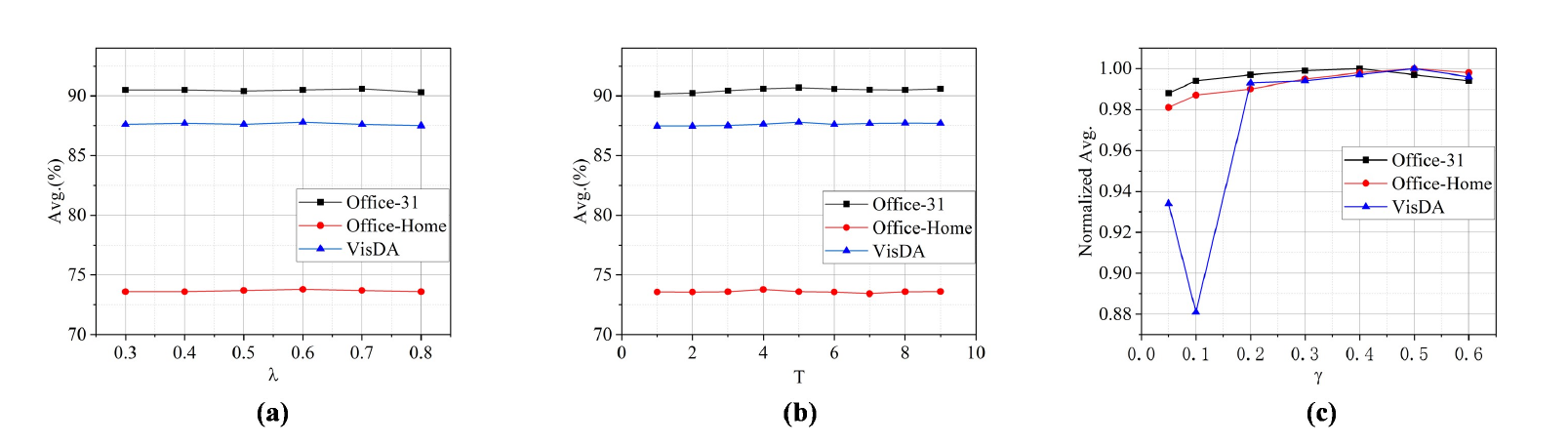}
\caption{Hyperparameter sensitivity analysis: (a) The effect of $\lambda$ on average accuracy (Avg.) across the three datasets. (b) The influence of the number of neighbors $T$ on Avg. across the three datasets. (c) The impact of different values of $\gamma$ across the three datasets. For clarity, the performance metric here is the normalized Avg., obtained by normalizing the Avg. of each dataset with respect to the best Avg. on that dataset. 
}
\label{fig:f4}
\end{figure*}

\begin{figure*}[t!]
	\centering
	\includegraphics[width=0.9\textwidth]{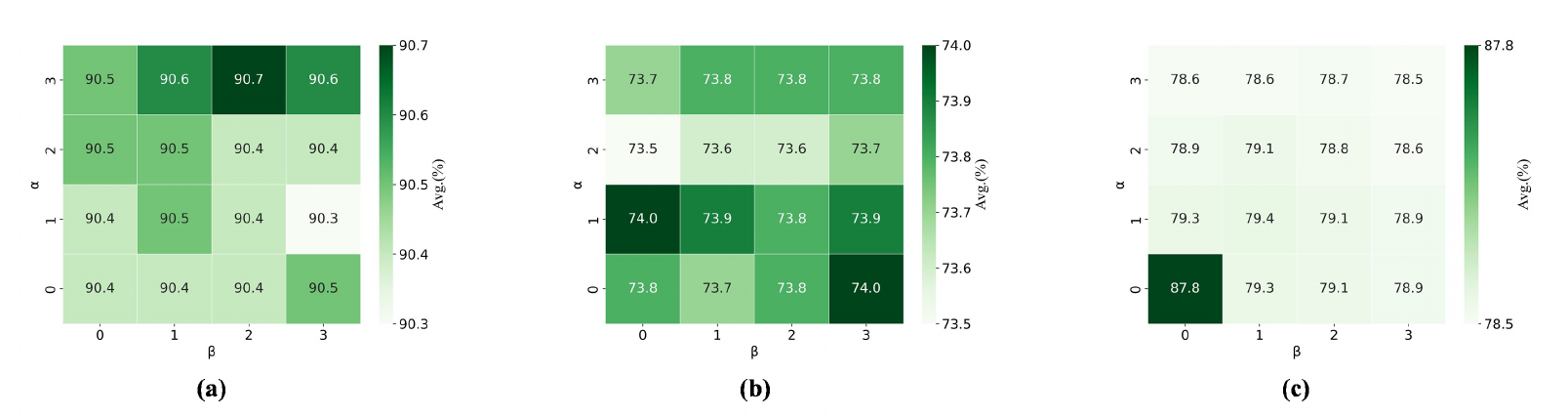}
	\caption{{{Hyperparameter sensitivity analysis: The impact of different combinations of $\alpha$ and $\beta$ values on Avg. in (a) Office-31, (b) Office-Home, and (c) VisDA.}}}
	\label{fig:ab}
\end{figure*}

\begin{figure*}[t!]
\centering
\includegraphics[width=1\textwidth]{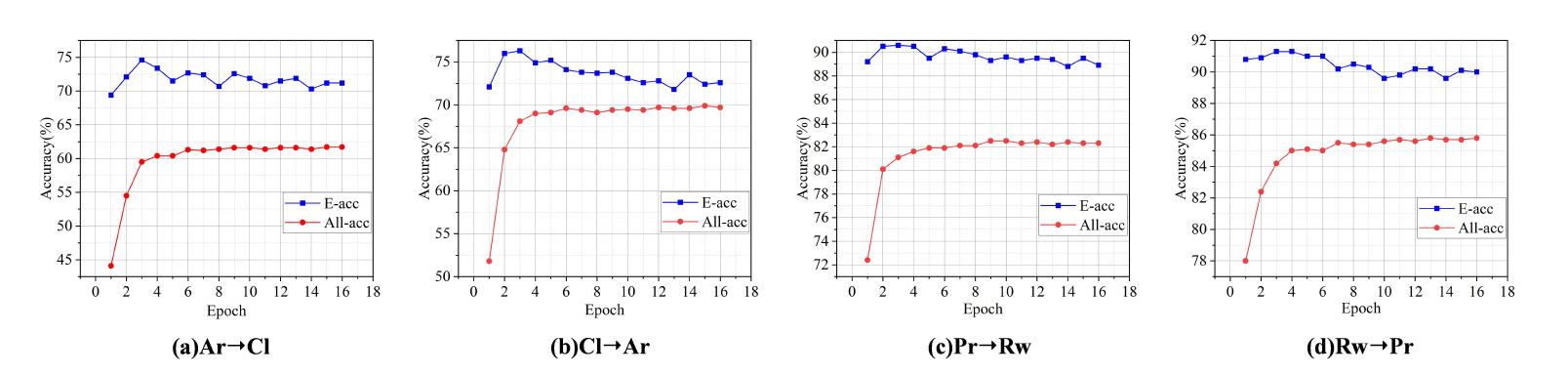}
\caption{Prediction accuracy for four Office-Home adaptation tasks. E-acc refers to the accuracy on selected easy samples, while All-acc represents all target samples. From the start, the accuracy of easy samples is significantly higher and remains stable, demonstrating the effectiveness of the selection strategy. All-acc rises quickly during the initial training phase, then gradually converges with minor oscillations, demonstrating rapid convergence and stability in the training of DPL.}
\label{fig:f5}
\end{figure*}

\subsection{Experimental Analysis}
\textbf{Feature visualization}.
For the adaptation tasks of A$\rightarrow$D on Office-31 and Rw$\rightarrow$Pr on Office-Home, we utilize t-SNE\cite{van2008visualizing} to visualize the feature space of the target domain extracted by both the source-only baseline and our proposed DPL. Fig.\ref{fig:f3} displays the visualizations, where each point represents a target sample, and distinct colors correspond to different categories. In (a) and (c), the chaotic sample distributions from the source pre-trained model highlight its inability to handle domain shifts effectively. Conversely, in (b) and (d), the samples extracted by our proposed DPL form well-separated clusters, with a clear distance between categories. This clustering shows the improved feature representation of our method, verifying its effectiveness in handling domain adaptation.
\begin{figure}[t!]
\centering
\includegraphics[width=0.48\textwidth]{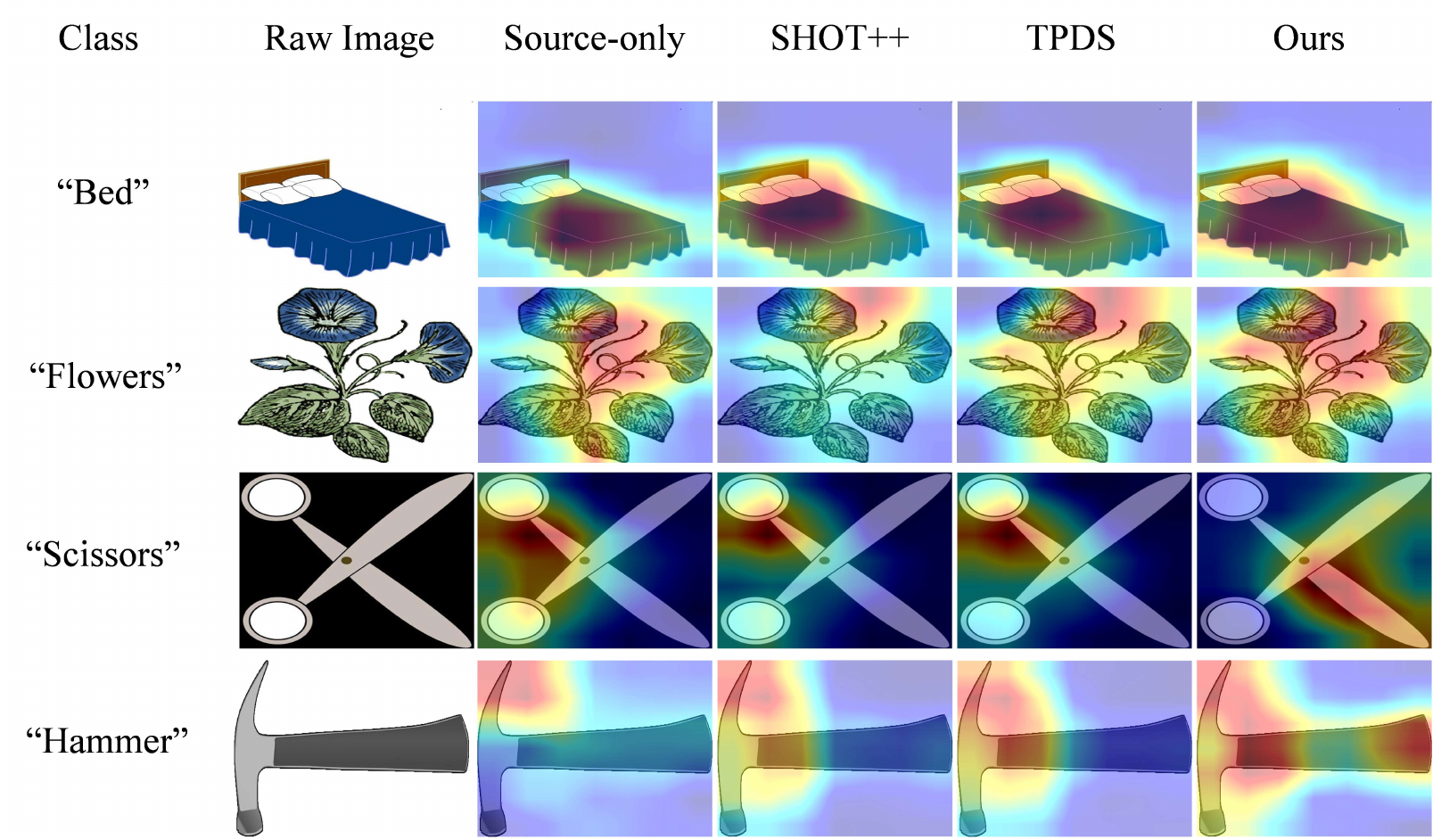}
\caption{Visualization of image regions vital for classification using Grad-CAM\cite{selvaraju2017grad}.
Hot colors indicate important regions.}
\label{fig:f6}
\end{figure}

\textbf{Hyperparameter sensitivity}.

As shown in Fig.\ref{fig:f4}, we perform a sensitivity analysis on the hyperparameters, including the coefficient $\lambda$ for the graph contrastive learning loss in stage one, the number of neighbors $T$, and the sample selection ratio $\gamma$. 
To ensure a fair comparison, we keep all parameters, except those being analyzed, consistent. 
The results in Fig.\ref{fig:f4} suggest low sensitivity of these hyperparameters, with the average accuracy (Avg.) showing robust performance across different values, which is highly advantageous in practical applications. 
For example, on any dataset, the Avg. difference across different $\lambda$ values is less than 0.3\%. Similarly, the effect of varying $T$ on the Avg. for different datasets is minimal. 
{Fig.\ref{fig:f4} (c) illustrates the impact of $\gamma$ on the performance of our method. For clarity, we normalize the Avg. for each dataset by the best Avg. of that dataset and use the normalized Avg. as the performance metric. Compared to the performance on Office-31 and Office-Home, the performance on VisDA is relatively unsatisfactory when the value of $\gamma$ is less than 0.2. As VisDA is more challenging than the other two benchmarks, we speculate that the easy-to-adapt subdomain of the target domain in VisDA requires more samples to achieve satisfactory results, suggesting that $\gamma$ should not be too small.}

The hyperparameters $\alpha$ and $\beta$, which balance the nuclear-norm loss $L_{nn}$ with other losses in stage one and stage two, also exhibit low sensitivity. 
The heatmaps in {Fig.\ref{fig:ab}} display the effect of different combinations of $\alpha$ and $\beta$ on the Avg. for Office-31, Office-Home, {and VisDA}, where darker colors indicate higher accuracy.
{$L_{nn}$ acts as a regularizer and encourages diversified model predictions. However, its practical effect is dependent on the scale of the dataset. 
As discussed in TPDS \cite{tang2024source}, when the dataset is small, limited data cannot accurately describe the probability distribution, and $L_{nn}$ helps promote balance in category predictions, thereby improving performance. In contrast, when the dataset scale increases, $L_{nn}$ may degrade performance. As shown in Fig.\ref{fig:ab}, as the dataset scale increases from Office-31 to Office-Home to VisDA, the values of $\alpha$ and $\beta$ decrease, with both reaching zero on VisDA, which supports the rationale behind the values we set for $\alpha$ and $\beta$ as discussed above.}

\textbf{Efficacy of the easy-to-adapt samples selection}.
To validate the efficacy of the easy-to-adapt sample selection, we track the model prediction accuracy on both the selected easy samples and all the target samples throughout the training process, denoted as 'E-acc' and 'All-acc,' respectively. As shown in Fig.\ref{fig:f5}, we conducted experiments on four adaptation tasks of Office-Home: (a) Ar$\rightarrow$Cl, (b) Cl$\rightarrow$Ar, (c) Pr$\rightarrow$Rw, and (d) Rw$\rightarrow$Pr. Despite minor fluctuations, the prediction accuracy of the selected easy samples remains stable and significantly higher than that of all the target samples from the start of training. This demonstrates the effectiveness of our easy sample selection strategy.

\textbf{Training convergence and stability}. As depicted by the All-acc curves in Fig.\ref{fig:f5}, we illustrate the training convergence and stability of our approach across four Office-Home adaptation tasks. The accuracy rises quickly in the initial training phase (before epoch 4), then gradually converges to the maximum value with minor oscillations. This shows that after rapid convergence, the training of DPL remains stable.

\textbf{Visualization of image regions vital for classification}. 
To demonstrate the superiority of our method, we utilize Grad-CAM\cite{selvaraju2017grad} to visualize the important regions of raw images for classification across different methods. As shown in Fig.\ref{fig:f6}, the red areas indicate regions that the models prioritize in predicting categories, while purple highlights less important regions. The first column displays the category names, followed by the raw images from Office-Home. The four rightmost columns present the visualizations corresponding to four methods. While Source-only, SHOT++, and TPDS struggle to localize important regions in some cases, our method consistently focuses on the discriminative areas, demonstrating a stronger discriminative ability.

\begin{table}[t]\renewcommand\arraystretch{1.3}
\caption{Ablation study. The left side of the table presents multiple combinations of different losses. The other side depicts the Avg.(\%) that each corresponding loss combination obtains on Office-31 and Office-Home.}
\centering
\scalebox{1}{
\begin{tabular}{ccccc|cc} 
\hline
 $L_{ce1}$ & $L_{gcl}$ & $L_{ce2}$ & $L_{cs}$ & $L_{icl}$ & Office-31 & Office-Home  \\ 
\hline
\usym{1F5F8}& \usym{2717} & \usym{2717} & \usym{2717} & \usym{2717} & 89.4 
& 71.1 \\
\hline 
 \usym{1F5F8}& \usym{1F5F8}&\usym{2717} &\usym{2717} &\usym{2717} & 89.6 
&71.3 \\
\hline
 \usym{1F5F8}& \usym{1F5F8}& \usym{1F5F8}&\usym{2717} &\usym{2717} & 90.0 
&73.3 \\
\hline
 \usym{1F5F8}& \usym{1F5F8}& \usym{1F5F8}& \usym{1F5F8}&\usym{2717} & 90.2 
&73.5 \\
\hline
\usym{1F5F8}& \usym{2717} & \usym{1F5F8}& \usym{1F5F8}& \usym{1F5F8}& 90.4&73.8\\
\hline
\usym{1F5F8}& \usym{2717} & \usym{1F5F8}& \usym{1F5F8}& \usym{2717} & 90.1&73.4\\
\hline
\usym{2717}& \usym{1F5F8}& \usym{1F5F8}& \usym{1F5F8}& \usym{1F5F8}& 89.9 &73.3 \\
\hline
\usym{1F5F8}& \usym{1F5F8}& \usym{1F5F8}& \usym{1F5F8}& \usym{1F5F8}& 90.7 &73.9 \\
\hline
\end{tabular}
}
\label{tab:t4}
\end{table}

\subsection{Ablation Study}
In this part, we analyze the contribution of different loss terms to the average accuracy that model achieves on Office-31 and Office-Home. The results are shown in Table \ref{tab:t4}, where ``\usym{1F5F8}" indicates the term is applied, and ``\usym{2717}" indicates it is not used during training.  

The last row of table \ref{tab:t4} represents DPL and its performance. In the first row, only $L_{ce1}$ is used, meaning only the uncertainty-aware self-training is applied. Compared to DPL, it suffers a significant performance drop on both datasets. When we incorporate the graph contrastive learning loss $L_{gcl}$ to align the easy-to-adapt and hard-to-adapt subdomains, the performance improves, verifying its effectiveness. 

Upon adding the training of stage two, we observe a remarkable improvement in model performance. Using pseudo labels for self-learning on easy-to-adapt samples alone leads to a 0.4\% and 2.0\% performance boost on Office-31 and Office-Home, respectively. This suggests that the pseudo labels for easy-to-adapt samples are reliable, a result of our accurate domain division. The incorporation of the consistency learning loss $L_{cs}$ further improves the model performance, indicating that it learns more discriminative features through dual augmentations of each easy-to-adapt sample. Finally, adding the instance contrastive learning (ICL) loss $L_{icl}$ allows the model to achieve optimal performance.

When both $L_{gcl}$ and $L_{icl}$ are ablated from DPL, the performance drops significantly. Using $L_{gcl}$ alone results in only a 0.1\% performance improvement. While applying $L_{icl}$ alone yields decent results, it is insufficient for the model to achieve optimal performance. However, when both $L_{gcl}$ and $L_{icl}$ are applied together, the model experiences a 0.6\% improvement on Office-31 and a 0.5\% improvement on Office-Home. The role of GCL is to adjust the feature distribution of hard-to-adapt samples, allowing ICL to have a more substantial impact in refining them. Together, these two components complement each other, driving the model towards optimal performance. {As shown in the penultimate row of table \ref{tab:t4}, ablating $L_{ce1}$ leads to a significant performance decline compared to the optimal results obtained by DPL on both benchmarks. This suggests that uncertainty-aware self-training is also crucial for achieving the best performance.}

\section{Conclusion}
In this work, our proposed Domain-Division based Progressive Learning method (DPL) effectively addresses the challenges of source-free domain adaptation by leveraging the inherent differences in adaptation difficulty within the target domain. By systematically dividing the target domain into easy-to-adapt and hard-to-adapt subdomains, DPL enhances classification accuracy through a two-stage process that incorporates uncertainty-aware self-training and tailored learning strategies. Our approach not only improves the model ability to learn from reliable samples quickly but also ensures that more complex, challenging samples are addressed with appropriate strategies. The extensive experimental results across various benchmarks confirm that DPL outperforms state-of-the-art methods, highlighting its robustness and adaptability in real-world scenarios. These findings underscore the potential of our method to facilitate effective adaptation while addressing growing privacy and portability concerns, making it a valuable contribution to the field of domain adaptation.

\bibliographystyle{IEEEtran}
\bibliography{ref}

\end{CJK}
\end{document}